\documentclass{article}

\usepackage{PRIMEarxiv}

\usepackage[utf8]{inputenc} 
\usepackage[T1]{fontenc}    
\usepackage{hyperref}       
\usepackage{url}            
\usepackage{booktabs}       
\usepackage{amsfonts}       
\usepackage{nicefrac}       
\usepackage{microtype}      
\usepackage{lipsum}
\usepackage{fancyhdr}       
\usepackage{graphicx}       
\graphicspath{{media/}}     

\usepackage{float}
\usepackage{comment}
\usepackage{todonotes}
\usepackage{ulem}
\usepackage{afterpage}

\title{Practical Digital Disguises:  Leveraging Face Swaps to Protect Patient Privacy
\thanks{This manuscript is a preprint that has not yet been peer-reviewed.}
}

\author{
    Ethan Wilson \\
    University of Florida \\
    \texttt{ethanwilson@ufl.edu}
    \And
    Frederick Shic \\
    Seattle Children's Research Institute \\
    University of Washington School of Medicine \\
    \texttt{fshic@uw.edu}
    \And
    Jenny Skytta \\
    Seattle Children's Research Institute \\
    \texttt{jenny.skytta@seattlechildrens.org}
    \And
    Eakta Jain \\
    University of Florida \\
    \texttt{ejain@cise.ufl.edu}
}

\begin{document}

\maketitle

\begin{abstract}

    With rapid advancements in image generation technology, face swapping for privacy protection has emerged as an active area of research. The ultimate benefit is improved access to video datasets, e.g. in healthcare settings. Recent literature has proposed deep network-based architectures to perform facial swaps and reported the associated reduction in facial recognition accuracy. However, there is not much reporting on how well these methods preserve the types of semantic information needed for the privatized videos to remain useful for their intended application. Our main contribution is a novel end-to-end face swapping pipeline for recorded videos of standardized assessments of autism symptoms in children. Through this design, we are the first to provide a methodology for assessing the privacy-utility trade-offs for the face swapping approach to patient privacy protection. Our methodology can show, for example, that current deep network based face swapping is bottle-necked by face detection in real world videos, and the extent to which gaze and expression information is preserved by face swaps relative to baseline privatization methods such as blurring.
    
\end{abstract}

\begin{figure}[h!]
    \centering
    \includegraphics[width=0.95\linewidth]{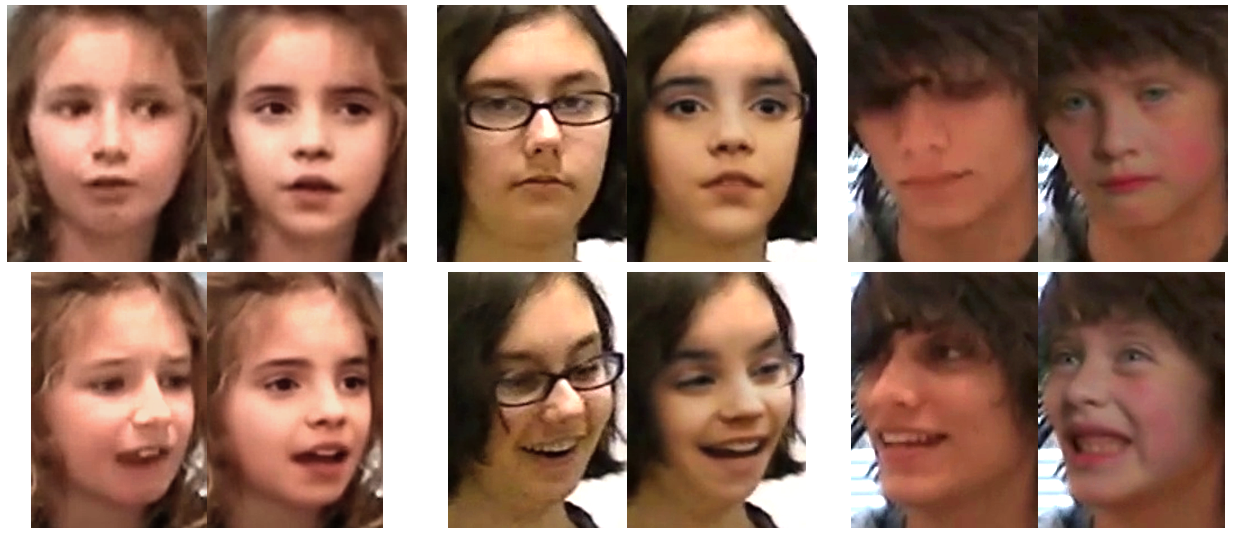}
    \caption{We present an end-to-end pipeline to face swap the faces of the child patients in videos of recorded autism screening sessions.}
\end{figure}


\section{Introduction}



Recent advances in generative technology~\cite{goodfellow:gan2014, radford:dcgan2016, karras:stylegan2019, larsen:autoencoding2016, iizuka:imagecomplete2017} have led to a huge advancement in face swapping technology~\cite{tolosana:dfandbeyond2020, nguyen:dldeepfake2021}\footnote{https://github.com/MarekKowalski/FaceSwap}\footnote{https://github.com/shaoanlu/faceswap-GAN}\footnote{https://github.com/iperov/DeepFaceLab}.  These algorithms, informally known as "deep fakes", are able to transpose a \textit{character face} onto the \textit{original face} in images or video clips seamlessly, fooling human viewers~\cite{korshunov:humandeepfakedetection2020}.

Although the technology is commonly used for impersonation~\cite{vaccari:dfanddisinfo2020, harris:lawpaper2018}, researchers are beginning to harness this tool and apply face swaps in ways that can benefit society~\cite{caporusso:dfforgood2021}.  There are emerging research topics assessing and improving upon the privacy protections that face swaps and deidentification guarantee~\cite{li:anonymousnet2019, gafni:deidentification2019, zhenzhong:de-identi2021}, and evaluating the added utility that face swapping can provide~\cite{zhu:medicaldf2020}.  

The existing literature fails to consider the properties of many real-world situations that require deidentification.  Unlike in proper face datasets, faces in real-world applications are typically low-resolution and distanced from the camera with a lot of movement.  For example, in clinical situations, the assessment is the focus; the recorded video is solely for reference.  In some settings, the subjects are children, whose faces are known to have a negative facial recognition performance when compared to adults~\cite{srinivas:facebias2019}.

To analyze these shortcomings, we designed an end-to-end system to deidentify patients in autism symptom assessments in children with autism spectrum disorder (ASD).  A novel contribution of our system is a quick and accurate user-in-the-loop facial annotation system that can guarantee accurate bounding boxes quickly, allowing face swaps in frames where automatic face detection algorithms fail.  We share our implementation details and design decisions with the intent to aid and assist further research pursuits in this field.  When designing the system, we considered privacy risks created by tele-health, remote training for clinicians, and dataset sharing for computational research.  

We analyze our processed sessions against the original clips and various degrees of blur to assess the privacy-utility trade-off that face swaps provide.  To generalize our findings, we report on three high-performing facial recognition algorithms.  We find that face swaps do provide a level of privacy proportional to the size of the set of reference faces being queried against.  We analyze facial landmarks, gaze direction, and expression classification to assess how well facial information is preserved.  

\section{Related Work}


Traditional methods, such as blurring or pixelization, have a significant drawback --- in order to provide a reliable level of privacy, all semantic information of the face is essentially destroyed~\cite{gross:integrating2006}.  Face swapping has been recognized as a powerful tool for face obfuscation that can act as an alternative.  This has sparked privacy-oriented research, as face swaps seem like a promising alternative to traditional privacy protections.    If a traditional obfuscation effect is weakened to where gaze direction, expression, and mouth movements can still be seen, privacy is compromised.  An addition, emerging deep learning methods, including full-body identity recognition~\cite{oh:faceless2016, Zhang:beyondfrontal2015} and models specialized to overcome the obfuscations~\cite{mcpherson:defeating2016}, are nearing the state of rendering traditional privacy methods obsolete.

Alternative methods have been explored in the past, dating back to 2005.  Fan et al. proposed a framework to automatically segment video clips into object and non-object regions~\cite{fan:ppvideo2005}.  This enabled less intrusive blurring and, when combined with contextual object classification, replacement of human subjects in video with 3D avatars.

To protect privacy in social media photos without obtruding on user experience~\cite{li:perceptions2018}, face in-painting approaches have been proposed.  Mahajan et al. provided the SwapItUp framework to swap in a reference face~\cite{mahajan:swapitup2017}.  Using the detected facial landmarks, pose and feature estimation is performed to query a nearest-matching face from a reference dataset.  This face is aligned, color matched and blended over the input.  
Sun et al. provided a GAN-based framework~\cite{sun:naturalheadinpainting2018} for facial in-painting in social media photos.  Given a partial or full-body image, facial landmarks are detected or generated using an adversarial auto-encoder network.  They then perform a structure-guided image generation.  Their method improves privacy protection against context-aware models, which are shown to fixate on the "fake" face.  When context-aware models predict blurred or masked faces, body cues are prioritized and the facial region is largely ignored.  However, these approaches lack control over the resulting identity or temporal consistency when applied to video.

Bailer and Winter proposed improvements to DCGAN~\cite{radford:dcgan2016} by applying portrait segmentation on the training data and adding face detection loss~\cite{bailer:improvingfacegen2019}.  Segmenting images to replace the background and adding facial structure awareness on top of the discriminator's unsupervised decision increased the face detection rate of the result.


Multiple approaches have opted to generate a fully synthetic face rather than face swapping with a reference identity.
AnonymousNet~\cite{li:anonymousnet2019} addresses the weaknesses of established deidentification techniques: lack of photo realism and an inability to balance privacy and usability.  By leveraging multiple privacy metrics, they generate a synthetic face satisfying k-anonymity and l-diversity privacy guarantees across 40 facial attributes.  They balance between privacy protections and generation through controllable hyperparameters, but their method tends to result in uncanny faces.  
Gafni et al. developed a fully automatic video deidentification pipeline able to be used in real-time~\cite{gafni:deidentification2019}.  This method provides privacy protections with apparent preservation in semantic features such as expressions and pose.  The method performs by driving the identity \textit{away} from a target face.  The method performs minor tweaks on the attributes of the face --- facial recognition systems are fooled, yet human viewers can likely still identify the individual.
Kuang et al. proposed DeIdGAN~\cite{kuang:de-identi2021}, improving on previous methods with a high deidentification rate and high image quality.  They first segment the image into semantic features, then impose k-anonymity on the mask's shape.  The segmented input is passed into a generator along with a style image, producing an image that follows the facial layout of the mask but produces the attributes of the passed in identity.  With novel identity-adversarial discriminators, the network teaches the generator to avoid resynthesizing the original identity.  They achieve reidentification rates equivalent to a powerful blur, yet preserve close to 100\% face detection rate and a high attribute preservation rate.  However, this method appears to misconstrue expressions, possibly due to the segmentation shape anonymity introduced before face generation.
These methods yield promising high-quality results, but each have flaws regarding face swapping without a reference identity.  AnonymousNet's fully generated, privatized face appears uncanny and Gafni et al.'s is perceptually similar to its source.  DeIdGAN and Gafni et al. have apparent difficulty preserving expressions.  AnonymousNet and DeIdGAN's incorporation of k-anonymity limit them to image datasets and no experimentation on video frames are shown.

Zhu et al. initiated the investigation into the real-world utility of face swaps as a viable privacy protection method for clinical setups~\cite{zhu:medicaldf2020}.  They performed face swaps on two examinations of the Unified Parkinson's Disease Rating Scale (UPDRS).  On their small dataset, they showed that the face swaps successfully deidentified the clips while preserving body keypoint invariability at a higher level than traditional face masking methods.  However, they discarded blurry and undetectable faces and did not explore more in-depth facial landmark preservation, which could be more meaningful for clinical diagnosis.

While generative models boast strong privacy guarantees, the literature does not support them being feasible in clinical settings.  Most methods work only on static images, and are not shown in in-the-wild situations.  Because emerging methods utilize face detection and recognition pipelines, it is unclear how they would perform on low-resolution face datasets that are relatively more difficult to reidentify in the first place.  Additionally, the generative architectures of these methods appear to only loosely preserve expression and gaze cues, which could have implications on human re-evaluation of the video for reference or training purposes.  Face swapping methods, which are shown to enirely fool human viewers~\cite{wohler_towards_2021, tahir_seeing_2021}, seem better suited towards this use case, but analysis into their privacy/utility trade-off is limited.

\section{System Design and Implementation}
We designed an end to end system which takes videos of autism screening sessions as input and creates a face swapped video as output. The constraints of the healthcare setting informed the design of the system presented below. Our system includes a novel user-in-the-loop annotation module that allows a lay user to mark bounding boxes in frames where automatic face detection algorithms fail~(Sec~\ref{sec:user-in-loop}).   

\subsection{Video Dataset: Constraints and Characteristics}
The patients in the video dataset are children ranging in age from 5-16 years old. Our dataset consists of 22 recorded administrations of the Autism Diagnostic Observation Schedule (ADOS), a gold-standard behavioral assessment instrument used in the diagnosis of ASD~\cite{lord1999ados, lord:ados22012}.  Sessions are about one hour in duration, with videos recorded at 60fps and 1280p resolution. There are two to three people in every session: the patient, the clinician, and optionally a parent/guardian. For the purposes of privatization, the patient is regarded as the \textit{key subject}.  The face of the privatizing identity, which will be swapped onto the key subject's likeness, is here-on referred to as the \textit{character face}.

Unlike talking head videos (referred to as medium close-ups in film), where the positioning of the speaker's face within the camera view and visual fidelity are the central concerns, these videos focus on capturing the interactive context of the clinical assessment.  As a result, the camera is installed in a corner of the assessment room and the faces are relatively low resolution compared to talking head videos. Importantly, the faces we get are the faces we get--we do not have the freedom to collect additional data. Finally, because our subjects are children, they frequently move around and present a variety of head poses and facial expressions. For example, it is a common occurrence for a child to spin around in their chair or run around the room, producing sequences with drastic movements and motion blur.

\subsection{Overall design}
Our system is a pipelined design.  This allows for multiple video clips to be processed simultaneously while maximizing the system resources on a single machine.  The GPU intensive model training can process one clip while less intensive stages in the pipeline run for other clips.  In clinics, after recording a session, the video could be processed locally, eliminating the security risks related to data transfer.

Our system can be divided into two main stages, each with a number of sub-stages.  These stages are:

\begin{enumerate}
    \item \textbf{Face Annotation.}  Results in a bounding box at each frame containing the key subject's face.
    \begin{itemize}
        \item \textbf{Automatic face detection.}  A face detection algorithm provides a list of bounding boxes for detected faces (of all identities present) across all frames. 
        \item \textbf{Manual corrections.}  A human annotator uses a GUI to perform multiple tasks:
        \begin{itemize}
            \item \textbf{(1.)} Label regions of time where the key subject is present with face in-frame.
            \item \textbf{(2.)} Mark identities not belonging to the key subject so that they can be ignored in later stages of the pipeline.
            \item \textbf{(3.)} Fill in gaps where automatic face detection failed by manually marking frames that should have bounding boxes for the key subject.
        \end{itemize}
        \item \textbf{(4.) Interpolation.}  Frames within the ground truth that do not contain bounding boxes are filled in by linearly interpolating nearest neighbors.  
    \end{itemize}
    \item \textbf{Face swap processing.}  The face annotation data is used alongside a character face dataset to produce a face swapped video result.
    \begin{itemize}
        \item \textbf{Face extraction.}  Face images are taken from the previously found bounding boxes and aligned.  This gives a dataset of face images for all frames containing the key subject's face.
        \item \textbf{Training.}  The key subject's dataset and a similarly structured character face's dataset are used to train a model that swaps the character's likeness onto the key subject's face images.
        \item \textbf{Deployment.}  Using the bounding boxes, the model is applied to every frame that contains the key subject's face.
    \end{itemize}
\end{enumerate}

We have made multiple design choices that consider the key subject.  There are simple modifications that could be made to instead address the privacy of multiple subjects.  For example, each key subject could be separately tagged in the same manual annotation session, then face swap processing could be run separately per-subject (or all at once, if you wish for the same character to be applied!).  To additionally protect privacy of non-key subjects, a blurring operation could be applied to the detected faces currently being filtered out.

\begin{figure}[p]
    \centering
    \includegraphics[height=0.9\textheight]{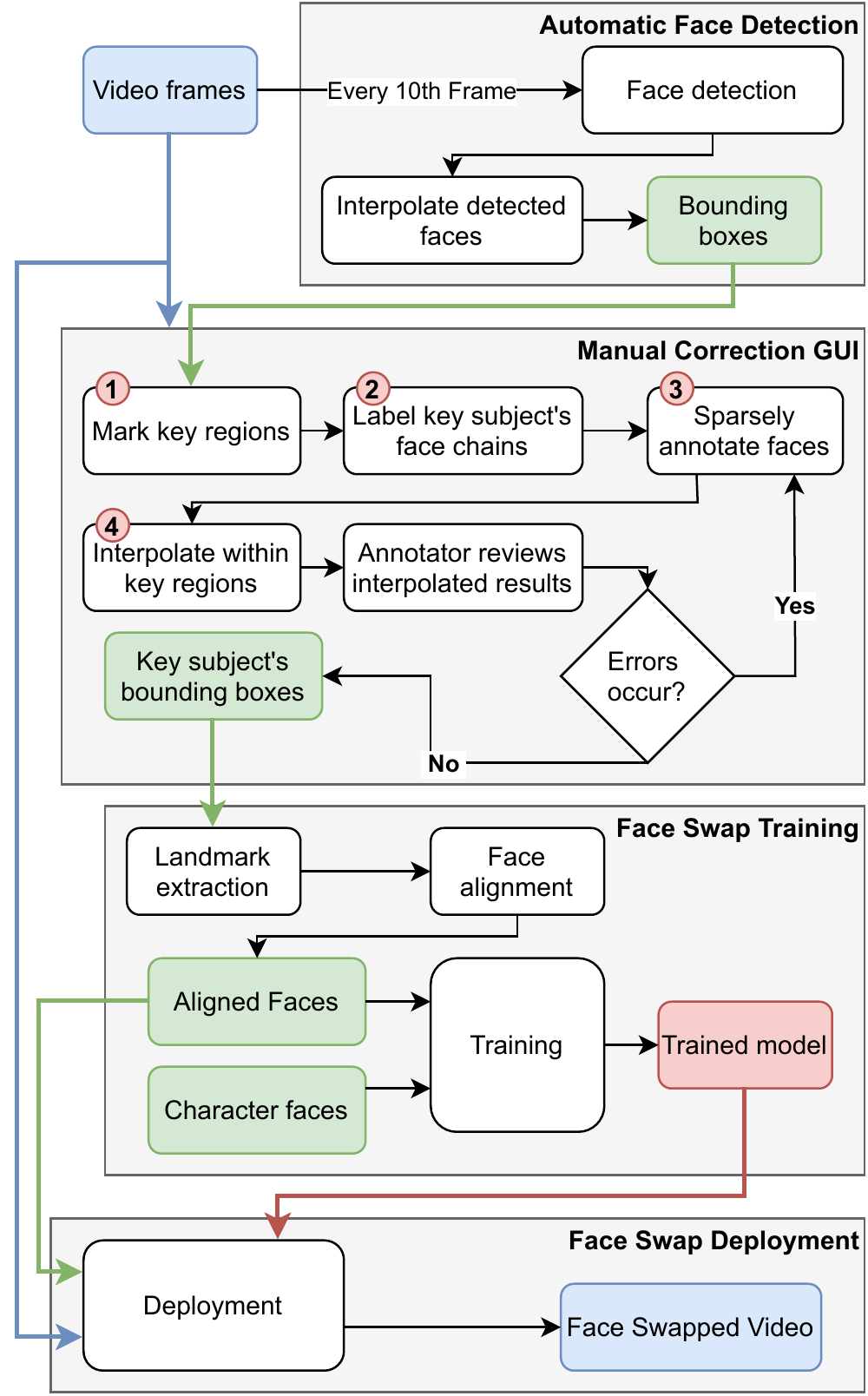}

    \caption{Overall system flow.  System is divided into automatic face detection, manual corrections via a graphical user interface, and face swap processing (subsectioned into training and deployment) implemented with DeepFaceLab's code-base~\cite{perov:deepfacelab2021}.}
    \label{fig:system_flow}
\end{figure}

\begin{table}[t] 
    \parbox{.55\linewidth}{
        \centering
        \begin{tabular}{@{}l c c c@{}}\toprule
           Detector & Time      & Detection Rate & Detection Rate \\ 
            & (Minutes) & (Single Face)  & (Multiple Faces)\\
           \midrule
           MTCNN & 1148.7 & 87.76\% & 33.93\% \\
           S3FD & 123.8 & 91.88\% & 62.57\% \\
           modified S3FD & 19.8 & 95.28\% & 86.47\% \\
           \bottomrule
        \end{tabular}
        \caption{Detection rates and runtime for the three face detectors considered, computed on a sample ADOS session consisting of two individuals.}
        \label{table:detections}
   }
   \hfill
   \parbox{.4\linewidth}{
        \centering
        \begin{tabular}{@{}l c c c@{}}\toprule
           Session & Automatic & Detection with \\ 
           & Detection & Annotations \\
           \midrule
           A & 57.29\% & 96.36\% \\
           B & 95.02\% & 98.43\% \\
           C & 73.11\% & 88.24\% \\
           D & 53.52\% & 86.83\% \\
           \bottomrule
        \end{tabular}
        \caption{A comparison of face detection rates for the key subject before and after manual processing for four exemplar video observation sessions.  These results illustrate a large variation in the effectiveness of automatic face detection, thus how much involvement is needed in manual correction.}
        \label{table:annotation}
    }
\end{table}

\subsection{Face Detection}

Face detection is the starting point of our system, and a high quality face detection result is necessary to extract accurate landmarks for alignment~\cite{minaee:facedetection2021} and ultimately create a high-quality face swapped result.  When choosing a face detection algorithm, we equally prioritized speed; a real-time or faster processing time is important for this system to be feasibly utilized in industry.  

\subsubsection{Comparison of face detection methods}

We analyzed two high performing, deep learning-based face detection systems and assessed their speed and performance: Multi-task Cascaded Convolutional Network (MTCNN)\footnote{https://pypi.org/project/mtcnn/}~\cite{Zhang:mtcnn2016} and Single Shot Scale-invariant Face Detector (S3FD)\footnote{https://github.com/iperov/DeepFaceLab/blob/master/facelib/S3FDExtractor.py}~\cite{Zhang:s3fd2017}.  Our S3FD implementation is bundled as the \textit{de facto} face detector in DeepFaceLab, used later in our pipeline.

We found that S3FD outperformed MTCNN on our dataset, then we further iterated on the S3FD's design to produce \textit{modified-S3FD}.  In S3FD image frames above a threshold size were scaled down to a target resolution before the algorithm proceeded.  Keeping the frame at a full resolution could improve detection rates but significantly slow down the process.  In \textit{modified-S3FD}, we run S3FD on the full resolution, but run the algorithm on every tenth frame (recall the video sessions are recorded in 60fps).  Then, we linearly interpolate the detected faces across the skipped frames.



\subsubsection{User in the loop to correct missed detection} \label{sec:user-in-loop}

We designed our system to overlay automatic face detection with a manual correction pass to lower the amount of labor required.  However, straightforward per-frame annotation still was not feasible.  For reference, Celeb-A is one of the largest annotated face datasets, with around 200,000 images~\cite{Liu:celeba2015}.  Our dataset of 22 sessions consists of 4.7 million image frames.  If we generously assume that automatic face detection marks 90\% of frames and that it takes a human annotator one second per frame, annotation would take 130 hours, or 6 hours per session.  This ballpark estimate does not consider false positives and the presence of more than one faces.  Therefore, it was imperative to design a system that would allow an annotator to quickly make corrections while remaining highly accurate.

\begin{figure}[t]
    \centering
    \includegraphics[width=0.85\linewidth]{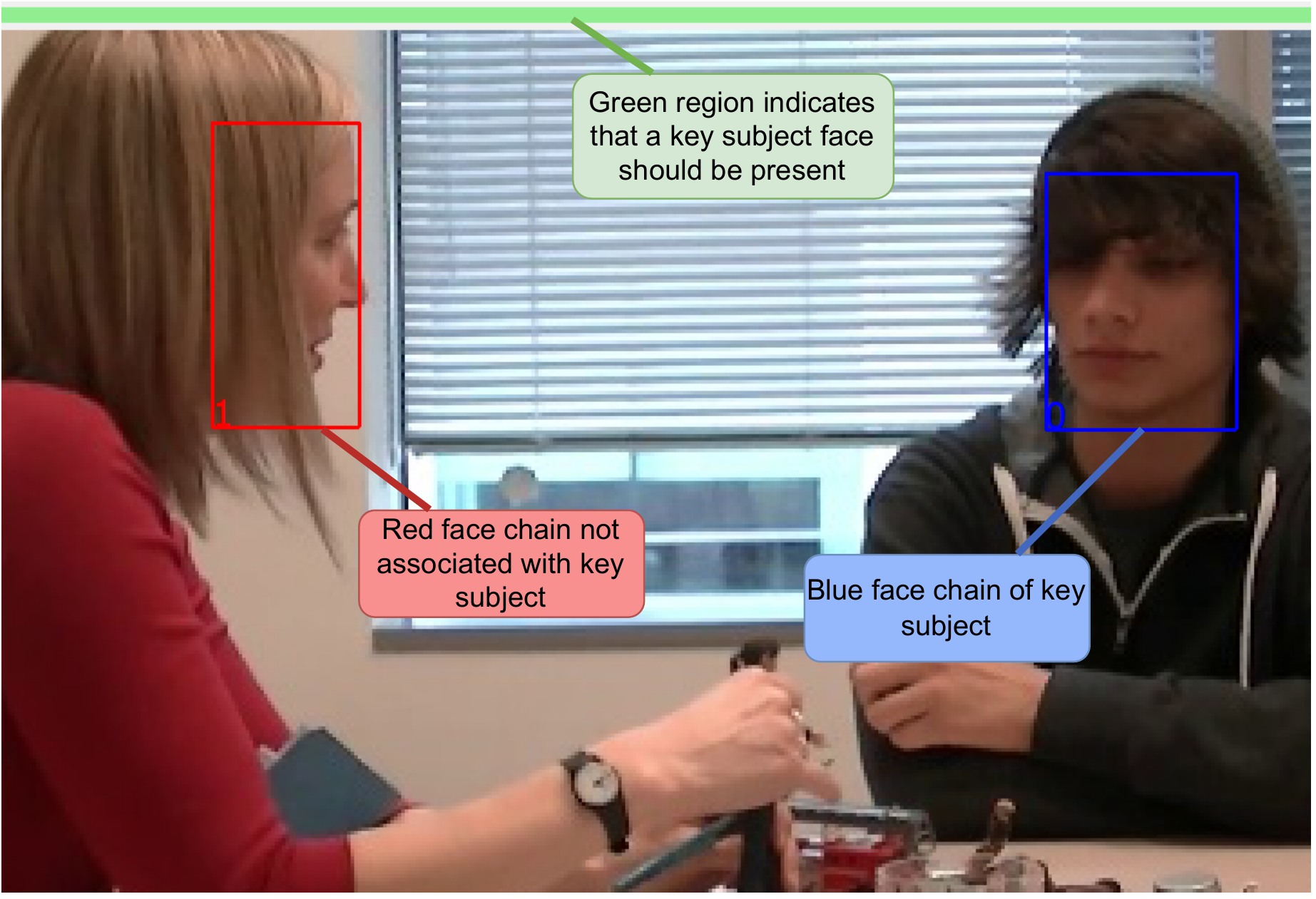}

    \caption{Manual annotation GUI being used to annotate the second pass of the proposed annotation system.}
    \label{fig:manual_gui}
\end{figure}

Our system is a user interface consisting of four passes, three manual and one algorithmic.  
A flow diagram of the full manual annotation process can be found in Figure~\ref{fig:system_flow} and an illustration of the GUI contents can be seen in Figure~\ref{fig:manual_gui}.

In the \textbf{first pass}, the annotator goes through the video and marks \textit{key frames} where the key subject's face enters or leaves the camera's view.  These key frames are important for further passes, as they allow the annotator to quickly jump to areas of interest.  

In the \textbf{second pass}, the annotator marks \textit{face chains} as belonging to the key subject or not.  A face chain is a continuous chain of faces for a particular subject, computed based on faces' pixel-wise distance in nearby frames.  The start and end frame of each face chain also becomes a key frame.  Annotators quickly jump between key frames and mark the face chains that belong to the key subject.

The \textbf{third pass} is used to supplement additional bounding boxes.  In regions where a large amount of time passes or the subject moves drastically, the annotator marks additional frames.  These frames are used in the fourth pass to aid with interpolation.  
In regions where the head slightly or does not move, the annotator can very sparsely mark frames and rely on interpolation.  However, if there is a large amount of movement, the annotator must mark the frames more tightly to get an accurate result.

The \textbf{fourth pass} is an algorithmic interpolation.  At this point, faces are labeled as belonging to the key subject or not, and the annotator has provided sufficient coverage of the key subject. The pass processes any frame that has been marked as containing the key subject but does not have facial information.  A new bounding box is created by interpolating the size and location of the nearest neighbor marked faces.

After the fourth pass, an annotator can view the result of their work and return to an earlier pass to make corrections if information was missed.  Once manual annotation has concluded and has been visually confirmed, the result can be considered the ground truth for face detection.  Regions containing the key subject were labeled in the first pass, and the fourth pass ensured that every frame within those regions has a bounding box.

The annotation system drastically reduced the workload compared to fully manual annotation.  Annotation typically took between 20-60 minutes per session, depending on session length and the starting accuracy of the automatic face detection algorithm.

\subsection{Face Swap Processing Details}

We use DeepFaceLab (DFL) as our face swapping framework~\cite{perov:deepfacelab2021}.  DFL is a popular open-source tool and has reported state-of-the-art results on the FaceForensics++ dataset~\cite{rossler:faceforensics++2019}.  
The overall stages are face extraction, training, and deployment.  We incorporate most of the DFL pipeline as-is, but make changes during the face extraction stage.

Rather than using DFL's face detector, we use the key subject's bounding boxes that were produced by our pipeline.  We then use the Face Alignment Network (FAN)\footnote{https://github.com/1adrianb/face-alignment} to perform extraction and alignment on the face data~\cite{bulat:fan2017}.  By passing in our manually tagged faces, we enable face swapping even when automatic face detection fails.

\subsubsection{Parameter choices}

DFL offers multiple architectures to be used in model training and deployment.  These architectures vary in quality, training times, and system resources required.  To fit our use case, processing many frames with faces that are quite small, we chose the dfHD architecture.  This architecture provided quality results with a processing rate that did not bottleneck the rest of the pipeline.  
Our architecture parameters were: \textit{resolution}: 128, \textit{face type}: whole face, \textit{auto-encoder dimensions:} 128, \textit{encoder dimensions:} 64, \textit{decoder dimensions}: 64, \textit{decoder mask dimensions}: 22, \textit{batch size}: 8.  Additionally, \textit{eyes and mouth priority}, \textit{random warping}, and \textit{flip faces} were all set to true.  When reimplementing this system for other datasets, different architectures and/or parameters can be chosen to best fit the use case.

We pretrained a model for 100,000 iterations using 100,000 face images extracted from six identities within our dataset.  For each session, we begin with the pretrained model and train 100,000 iterations further on the key subject paired with a character face.


\section{Evaluation}


In this section, we provide a methodology for assessing the privacy-utility tradeoff for the face swapping approach to patient privacy protection.  We compute the recognition accuracy of multiple high performing face detection algorithms on the privatized videos, as well as computing three sets of metrics that capture the fidelity of gaze and expression information, which are critical cues for autism screening.

In the following analysis, we consider 1/10th of the total number of frames contained in our recorded sessions.  This reduces the scale of our analysis from 3.2 million faces to approximately 320,000.  

\subsection{Recognition Accuracy (Privacy Protection)}

We report face detection accuracy in two scenarios:

\begin{itemize}
    \item Small-scale representation using hand-selected images.
    \item Large representation using frames from across the full session.
\end{itemize}

These scenarios simulate two forms of attacks.  In the first, an adversary with access to a clinical recording tries to identify a specific individual and has the time and resources to hand select ideal frames.  This bypasses the difficult nature of these videos where the patient's face can often be obscured or unclear by selecting high quality samples.  The second scenario models an automated attack where the adversary would process the frames of the session and aggregate the face detection results to make a final prediction as to the patient's identity.

\begin{figure}[b]
    \centering
    \includegraphics[width=0.9\linewidth]{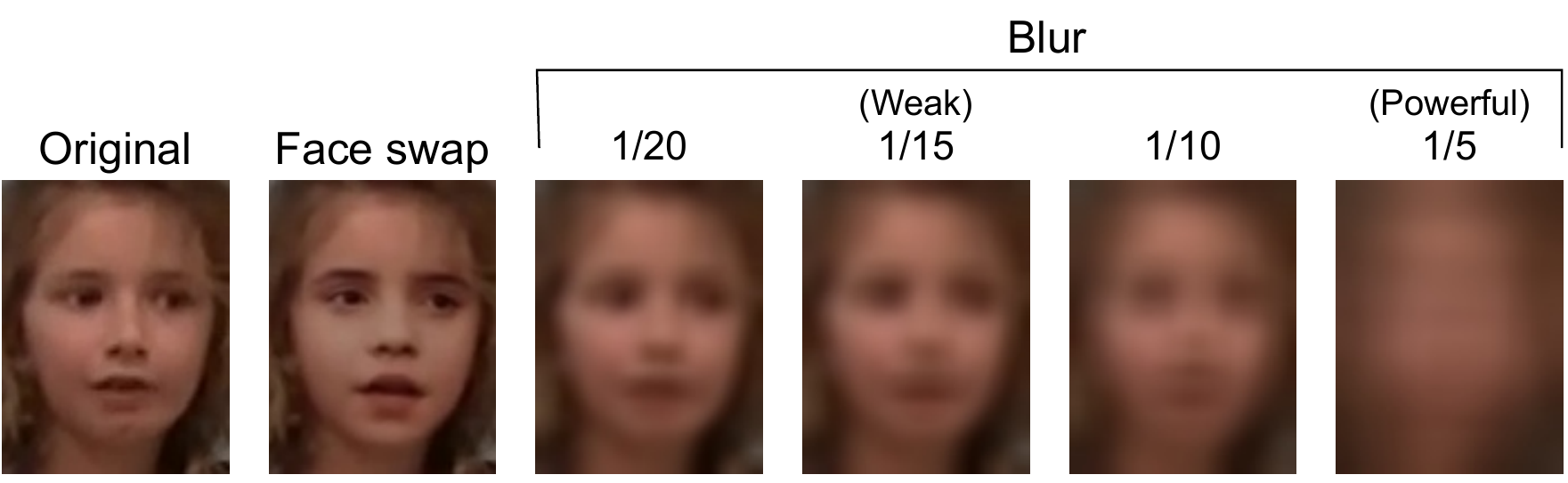}
    
    \caption{
    Illustration of blur intensities.  Blur intensities titled "Weak" and "Powerful" are used in further utility analysis.  Weak blur provides a roughly equivalent privacy protection to face swapping on our data.  Powerful blur provides strong privacy guarantees but fully degrades the face.}
    \label{fig:blur_visual}
\end{figure}

There are many face detection algorithms available, and we assume that an adversary can use any (or multiple) algorithms in a reidentification attack.  To address this, we report our results on three high-performing face recognition algorithms: Facenet~\cite{schroff:facenet2015}, DeepFace~\cite{taigman:deepface2014}, and ArcFace~\cite{deng:arcface2019}\footnote{Implementation of facial recognition systems used: https://github.com/serengil/deepface}. 
  
\begin{table}[t]
    \centering
    \begin{tabular}{@{}r | c c | c c c c@{}}
        & & & \multicolumn{4}{c}{\textbf{Blur Scale}} \\ 
        & Original & Deepfakes & 1/5 & 1/10 & 1/15 & 1/20\\ 
        \hline
        \textbf{Facenet} &&&&&\\
        Accuracy & 94.6\% & 0.65\% & 0.1\% & 0.1\% & 6.33\% & 45.46\% \\
        Median & 1 & 153 & 490 & 473 & 159 & 2 \\
        Mean & 2.16 & 239.55 & 493 & 478.28 & 283 & 54.67 \\
        $\pm$SD & $\pm$19.9 & $\pm$242 & $\pm$289 & $\pm$291 & $\pm$282 & $\pm$142 \\
        \hline
        \textbf{DeepFace} &&&&&\\
        Accuracy & 78.69\% & 3.61\% & 1\% & 12.64\% & 45.13\% & 64.51\% \\
        Median & 1 & 108 & 360 & 70 & 2 & 1 \\
        Mean & 24.76 & 220.29 & 400.82 & 203.09 & 78.55 & 45.6 \\
        $\pm$SD & $\pm$106 & $\pm$255 & $\pm$293 & $\pm$265 & $\pm$179.3 & $\pm$141 \\
        \hline
        \textbf{ArcFace} &&&&&\\
        Accuracy & 96.71\% & 0.72\% & 0.1\% & 0.58\% & 70.97\% & 92.5\% \\
        Median & 1 & 120 & 482 & 75 & 1 & 1 \\
        Mean & 2.42 & 223.17 & 487.99 & 164.32 & 6.02 & 2.91 \\
        $\pm$SD & $\pm$29.5 & $\pm$246 & $\pm$288 & $\pm$206 & $\pm$38.4 & $\pm$32.2 \\
    \end{tabular}
    \caption{Face recognition results computed using uniformly selected frames from the FaceForensics++ dataset.  Reference face dataset consisted of 1000 individuals.  Results are shown for the original and Deepfakes stimuli, as well as blurred results at various scales.}
    \label{table:faceforensics}
\end{table}

\begin{figure}[b]
    \centering
    \begin{minipage}{.49\linewidth}
        \includegraphics[width=\linewidth]{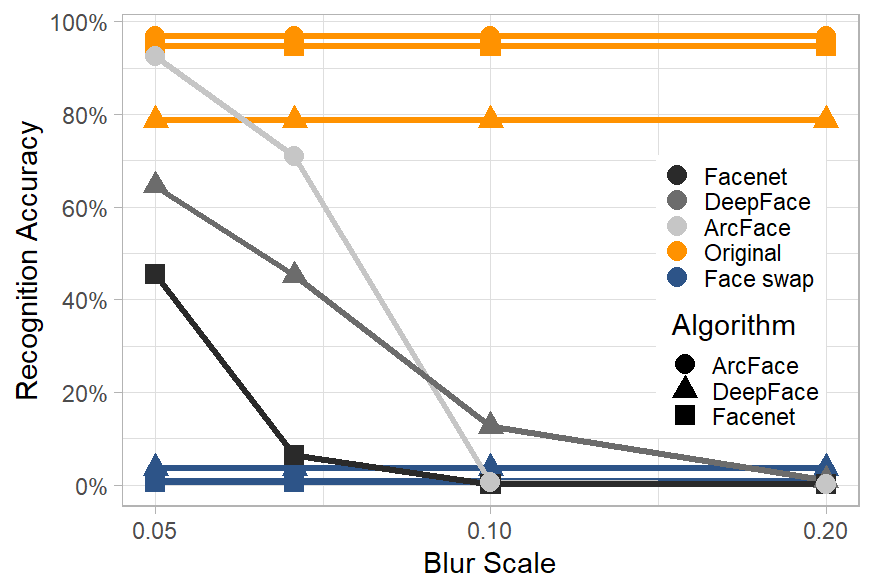}
        \caption{Recognition accuracy at varying blur scales.  Original and face swapped face detection rates are displayed for reference.  Reported results are for the FaceForensics++ dataset (1000 identities).}
        \label{fig:blur_intensity}
    \end{minipage}
    \hfill
    \begin{minipage}{.49\linewidth}
        \includegraphics[width=\linewidth]{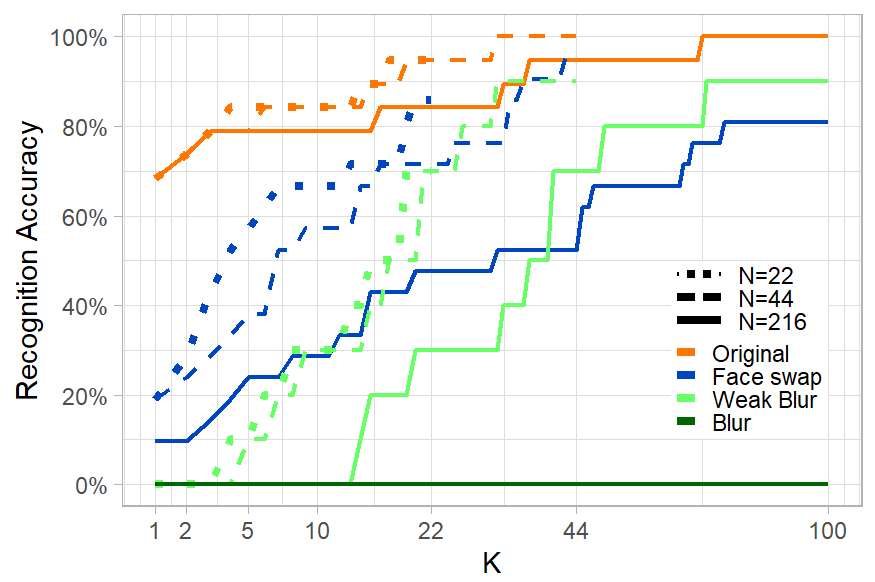}
        \caption{ArcFace facial recognition accuracy is displayed of a hand-selected subset of video frames.  Accuracy indicates that the true identity was within K closest matches.  As number of reference identities N increases, recognition accuracy falls off.}
        \label{fig:setup3}
    \end{minipage}
\end{figure}

We compare our results against a baseline privatization method: blur.  We implement a scaled box filter smoothing operation with a blur kernel intensity equal to $ceil((width + height) * scale)$.  Table~\ref{table:faceforensics} and Figure~\ref{fig:blur_intensity} show the impact that different blur scales have on facial recognition accuracy. 
 For further experiments, we compare against a powerful blur set at $scale=1/5$ and a weak blur set at $scale=1/15$.  This weak blur has a much higher recognition accuracy on FaceForensics++ data but comparable privacy protections to face swapping on our dataset. 
 
Recognition accuracy is defined as the percentage of correct identifications in the test set. A computed identity is correct if the true identity of an individual is within \textit{K} closest results of feature-wise distance vectors computed by the recognition algorithm.  Thus, when $K=1$, the correct identity has been identified as the closest match.  For more relaxed values of K, the algorithm has identified the correct identity \textit{alongside} $[K-1]$ reference images of potentially false identities. In Figure~\ref{fig:setup3}, we report recognition accuracy at different values of \textit{K}.  

\begin{table}[t]
    \centering
    \begin{tabular}{@{}l r | c c c@{}}
        && Original & Face Swap & Blur (S=1/5) \\ 
        \hline
        \textbf{Facenet} & K=1 & 39.33\% & 16.03\% & 0\% \\
        & K=2 & 50.85\% & 26.89\% & 0\% \\
        & K=5 & 65.82\% & 50.42\% & 0.04\% \\
        & K=10 & 78.25\% & 74.81\% & 0.13\% \\
        Identity & Median & 2 & 9 & 20 \\
        Ranking: & Mean$\pm$SD & 6.63$\pm$8.6 & 12.65$\pm$13.7 & 19.43$\pm$9.7 \\
        \hline
        \textbf{Deepface} & K=1 & 69.24\% & 28\% & 0.05\% \\
        & K=2 & 76.59\% & 43.58\% & 0.08\% \\
        & K=5 & 86.65\% & 66.17\% & 0.15\% \\
        & K=10 & 93.22\% & 87.24\% & 0.25\% \\
        Identity & Median & 1 & 6 & 11 \\
        Ranking: & Mean$\pm$SD & 3.22$\pm$6.2 & 10.99$\pm$12.6 & 15.28$\pm$16.4 \\
        \hline
        \textbf{ArcFace} & K=1 & 63.03\% & 18.51\% & 0\% \\
        & K=2 & 67.7\% & 30.55\% & 0.06\% \\
        & K=5 & 73.06\% & 52.34\% & 0.08\% \\
        & K=10 & 78.03\% & 73.95\% & 0.2\% \\
        Identity & Median & 1 & 9 & 23 \\
        Ranking: & Mean$\pm$SD & 6.81$\pm$14.5 & 15.72$\pm$18.7 & 30.21$\pm$31.5 \\
    \end{tabular}
    \caption{Face recognition results computed across all clinical sessions processed.  Reference face dataset consisted of 110 images of 22 individuals.  Detection accuracy is reported at ranks K=1, 2, 5, 10.  Median and mean identity rankings are reported as the ranking of the first true-positive reference image.}
    \label{table:setup4}
\end{table}



For the first adversarial scenario, we hand selected five frames from each session and cropped the images to the key face region.  We split these frames into query and reference datasets consisting of two and three frames per subject, respectively. The recognition accuracy at different levels of \textit{K} is shown as dotted lines in Figure~\ref{fig:setup3} (N=22). We augment the test set by adding more identities as the reference databases in the defined threat scenario would typically be much larger than our 22 identities. We include additional identities, children aged 5-15, procured from the IMDB-Face dataset~\cite{wang:devil2018}. Recognition accuracy is shown as dashed lines (N=44) and solid lines (N=216). In all cases, the face swapping approach reduces recognition accuracy relative to the original faces.  The powerful blur rendered the faces undetectable, yielding a 0\% recognition rate.

For the second scenario, we validate the identity of every tenth frame of the recorded video sessions using five frames per identity as a reference set.  We report these results in Table~\ref{table:setup4}. In addition to the recognition accuracy at different \textit{K}, we also report the \textit{identity ranking}: the index of the true face's facial feature distance within the reference dataset, as reported by~\cite{gafni:deidentification2019}.    To compare our results with existing benchmarks, we also report the mean index of the original and DeepFakes\footnote{https://github.com/deepfakes/faceswap} stimuli contained within the FaceForensics++ dataset~\cite{rossler:faceforensics++2019}, processing ten uniformly spaced frames from each video.

Our results indicate that face swapping does provide a measure of deidentification for the subject's face. 
The large standard deviations in the identity rankings indicate that facial recognition accuracy is highly variable for in-the-wild settings, both before and after manipulation.  Face swapping was able to consistently double the identity ranking across multiple facial recognition algorithms.

Our results align with the results computed on the FaceForensics++ benchmark, but with a higher recognition rate for the face swapped images (15.69\%/27.75\%/18.37\% versus 0.65\%/3.61\%/0.72\%).  We hypothesize this variation is influenced by reference dataset size; the 22 individual ADOS set is much lower than the 1000 individual set.  This indication can be seen in Figure~\ref{fig:setup3}, where a larger reference dataset more sharply decreased recognition accuracy for the face swaps than for the original faces.  Furthermore, the FaceForensics++ reference set always contains the identity that is being used as the character face in the face swap, guaranteeing a closely-resembling result that is not the original identity.

\begin{figure}[h]
    \centering
    \includegraphics[width=0.75\linewidth]{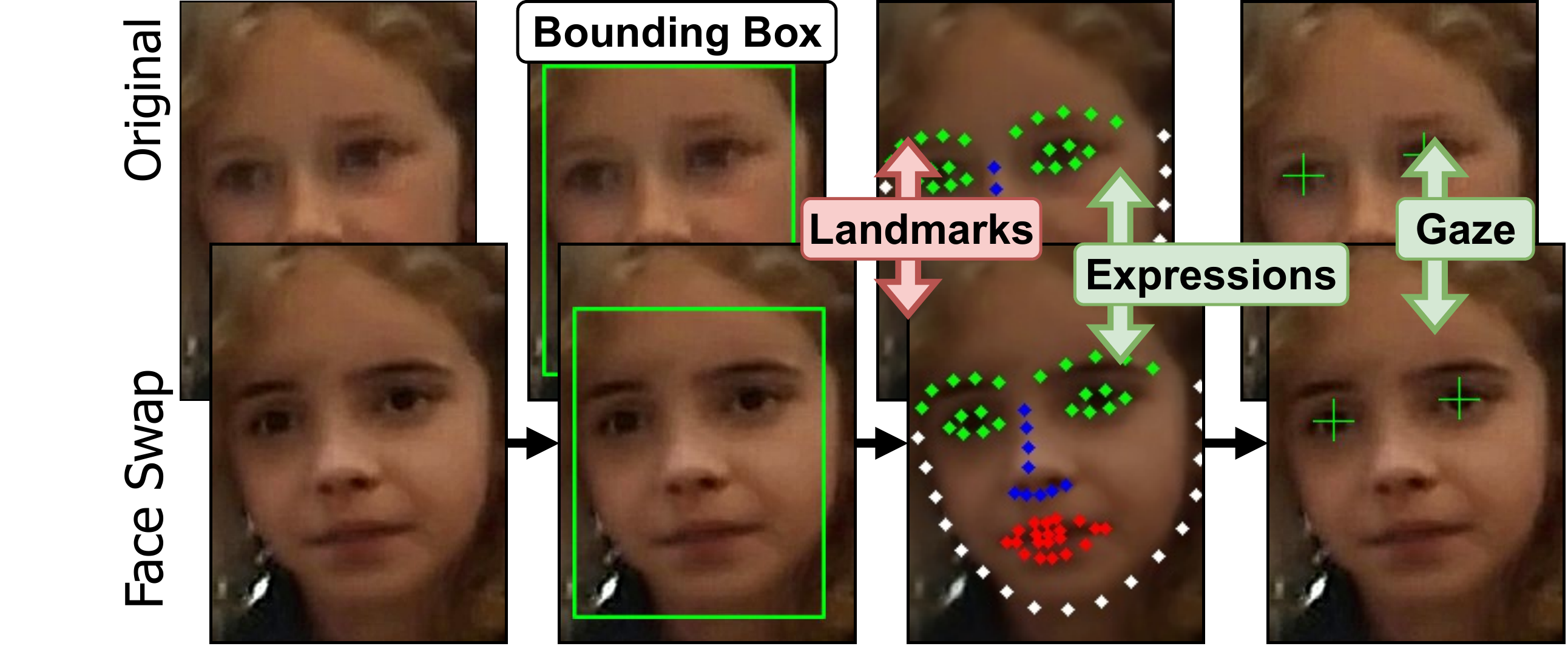}

    \caption{Utility analysis pipeline.  Red arrow indicates value comparison and green arrows indicate classification.  Once landmarks are computed, expressions are estimated and the original and face swap are compared.  Using predictions built off of landmarks, gaze classifications (left/center/right; up/center/down) are compared.}
    \label{fig:utility_flow}
\end{figure}

\subsection{Landmarks Estimation}

Landmark points are the basis for further processing of human faces, including facial expression classification and gaze direction estimation.  Euclidean distance is a widely used metric when using landmarks for face recognition and expression detection~\cite{vezetti:landmarkmeasures2012}.  Because the pixel sizes of faces in our dataset are not standardized, we report the Euclidean distance normalized within the bounding box of the detected face, shown in Equation~\ref{eqn:landmarks}.  

\begin{equation}
    D_{norm} = \frac{1}{68} \sum_{i=1}^{68} \sqrt{(\frac{O_i.x - DF_i.x}{Face_w})^2 + (\frac{O_i.y - DF_i.y}{Face_h})^2}
    \label{eqn:landmarks}
\end{equation}

\begin{table}[t]
    \centering
    \begin{tabular}{@{}r | c | c | c c c@{}}
    & Number   & Normalized & & & \\
    & of Faces & Euclidean  & \multicolumn{3}{c}{\textbf{Per-feature Distance}} \\
    & Compared & Distance   & Eyes & Nose & Mouth \\
    \hline
    Face swap & 121949 & 0.0318 & 0.0245 & 0.0270 & 0.0332 \\
    Weak Blur & 49421 & 0.0375 & 0.0284 & 0.0319 & 0.0373 \\
    Strong Blur & 992 & 0.1681 & 0.1682 & 0.2111 & 0.1850
    \end{tabular}
    \caption{Average normalized Euclidean distance between the landmark points of original and swapped faces.}
    \label{table:landmarks}
\end{table}

$D_{norm}$ is averaged over all 68 key points computed by the Dlib facial landmark predictor~\cite{king:dlib2009, kazemi:facealignment2014}.  The original face's points $O$ and the deepfake points $DF$ are scaled between 0 and 1 within the bounds of the $Face$ region.  

We report the distance between landmark estimation on the original faces with the same estimation on the face swaps and on the blurred faces in Table~\ref{table:landmarks}.  We also report the distance separated by key regions (eyes, nose, mouth) by averaging across select landmark points.  We find that face swaps' landmarks are more true-to-original for eyes and nose than those of the mouth.  Face swap and weak blur have relatively similar landmark estimations, yet far less blurred faces were able to be detected for analysis. 

\subsection{Gaze Direction Estimation}

\begin{table}[ht]
    \centering
    \begin{tabular}{@{}l c c r c c@{}}\toprule
       Stimuli & Total & \% Over   & \% Gaze  & Accuracy \\ 
               & Faces & Threshold & Detected & \\
       \toprule
       Original & 53915 & 82.21\% & 38.26\% & 100\% \\
       \midrule
       Face swap & \textquotedbl & \textquotedbl & 54.23\% & 68.98\% \\
       Weak Blur & \textquotedbl & \textquotedbl & 0\% & / \\
       Strong Blur & \textquotedbl & \textquotedbl & 0\% & / \\
       \bottomrule
    \end{tabular}
    \caption{Gaze detection rate and accuracy results.  We report the percentage of frames considered after threshold application and the percentage of those thresholded frames where gaze was detected and correctly classified.}
    \label{table:gaze}
\end{table}

This section evaluates how well the gaze direction after face swapping matches the gaze direction in the original faces. In the metrics reported, gaze direction (estimated by the GazeTracking network\footnote{https://github.com/antoinelame/GazeTracking}) of the original faces is considered ground truth.  To assess how well gaze is preserved, we implement a classification approach, labeling gaze as \{left, center, right\} and \{up, center, down\}, for a total of nine classes.  


Because the size of faces can be very small, we employed a threshold-based cut off such that only faces of $min(width, height)$ $>=$ $56$ pixels are processed.  Sessions where the GazeTracking network failed to detect a valid gaze direction in at least 10\% of the frames (from lack of contrast or non-viable poses) were discarded.
Of the 12 sessions considered (N=53915 frames), once the threshold was applied, M=44326 frames remained.  We report the percentage of valid gaze samples and the accuracy of the gaze direction in Table~\ref{table:gaze}.  As we can see, even weak blurring fully degrades gaze information and no utility can be preserved.  However, face swapping is able to detect gaze in 54.23\% of frames and predict with 68.98\% classification accuracy.  Interestingly, many frames that did not originally have detectable gaze were able to be classified after face swapping.  This indicates that face swaps could impose an average gaze direction on frames that are originally occluded. 

\subsection{Expression Detection}

\begin{figure}[ht]
    \centering
    \includegraphics[width=0.95\linewidth]{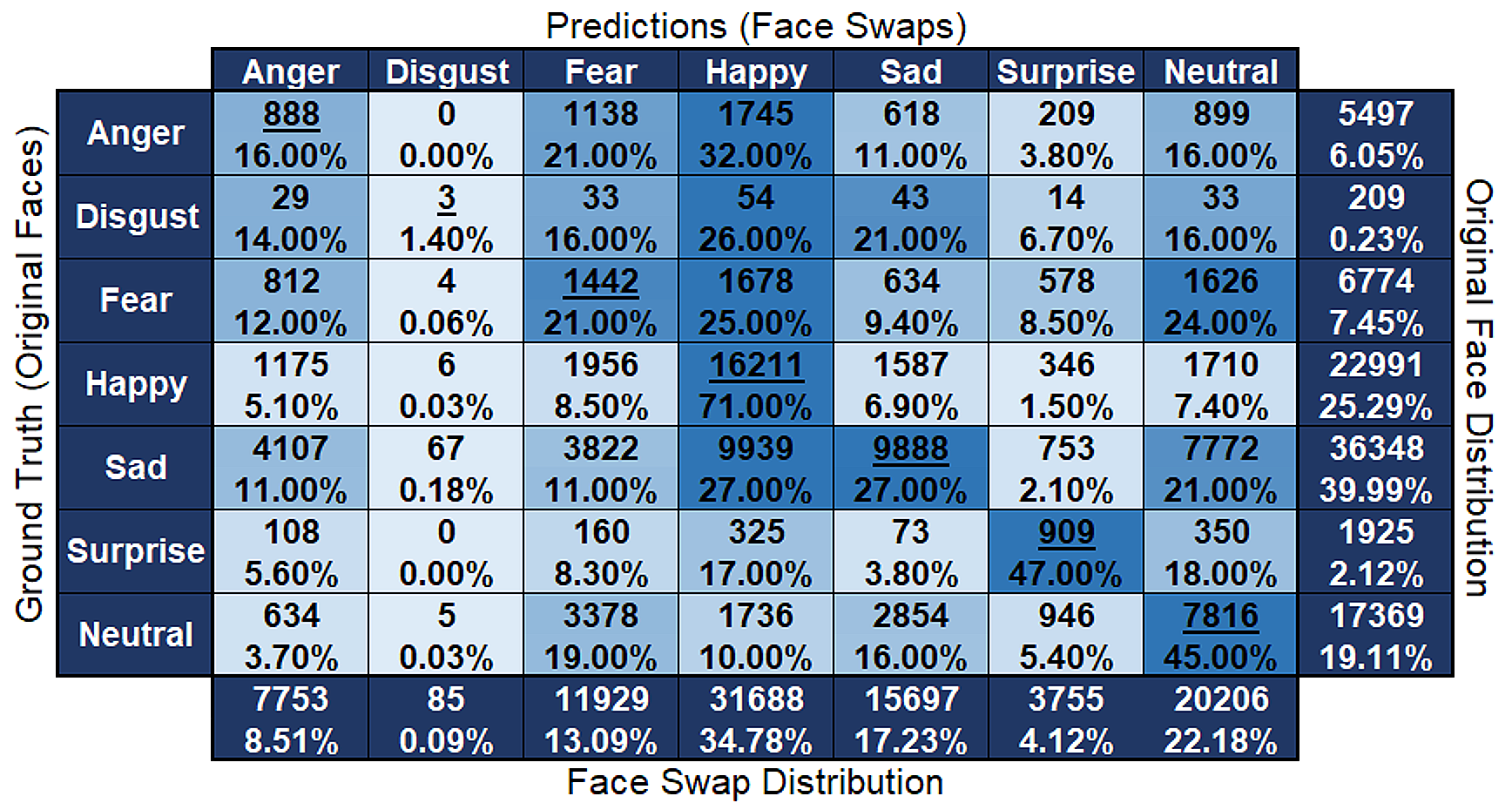}

    \caption{Confusion matrix illustrating how well the face swapped result matched the algorithmically classified expression on the original face (ground truth).}
    \label{fig:expressions}
\end{figure}


To assess expressions, we employ the facial expression recognition network (FER\footnote{https://github.com/justinshenk/fer})~\cite{arriaga:emotiongender2017, goodfellow:fer2013}.  We consider the computed expressions produced by the network for the original faces to be the ground truth.  The face swapped faces (N=91113 comparisons) were classified as having the same expression as the original for 40.78\% of the frames.  Weak blurred faces (N=62396 comparisons) were correctly classified only 27.27\% of the time.  A confusion matrix illustrating the face swap classifications can be found in Figure~\ref{fig:expressions}. 

While the expression classification rate of the face swaps is not ideal, the distribution of expressions is rather similar between face swaps and original faces.  Original faces are distributed from most concentrated to least as \{Sad, Happy, Neutral, Fear, Anger, Surprise, Disgust\}.  The face swap distribution is \{Happy, Neutral, Sad, Fear, Anger, Surprise, Disgust\}, following the overall ground truth distribution order other than the \textit{Sad} class.

\section{Conclusions}

Our findings shed light on some of the challenges in real-world scenarios that have not been considered in prior research.  When implemented on high-quality data, face swapping algorithms may be able to preserve expression and gaze information.  However, results are far less reliable on challenging data, which could lead to misinterpretations of face swapped video.  In use cases such as clinical assessments for autism screening, where atypical gaze or eye contact are important indicators\cite{american2013diagnostic}, it is important to consider the level of privacy offered by a privatization approach in conjunction with utility, i.e., the retention of those cues that are needed by the clinician to make a diagnosis. 

\section*{Acknowledgments}
This work was was supported by the National Institute of Mental Health through award R21MH123997. We thank Dr. Sara Jane Webb and Megha Santosh for their help in provision of media-release approved ADOS videos for use as exemplars in this work.

\bibliographystyle{unsrt}  
\bibliography{mainbibliography}  

\appendix

\begin{figure}[h]
    \centering
    \includegraphics[width=0.85\linewidth]{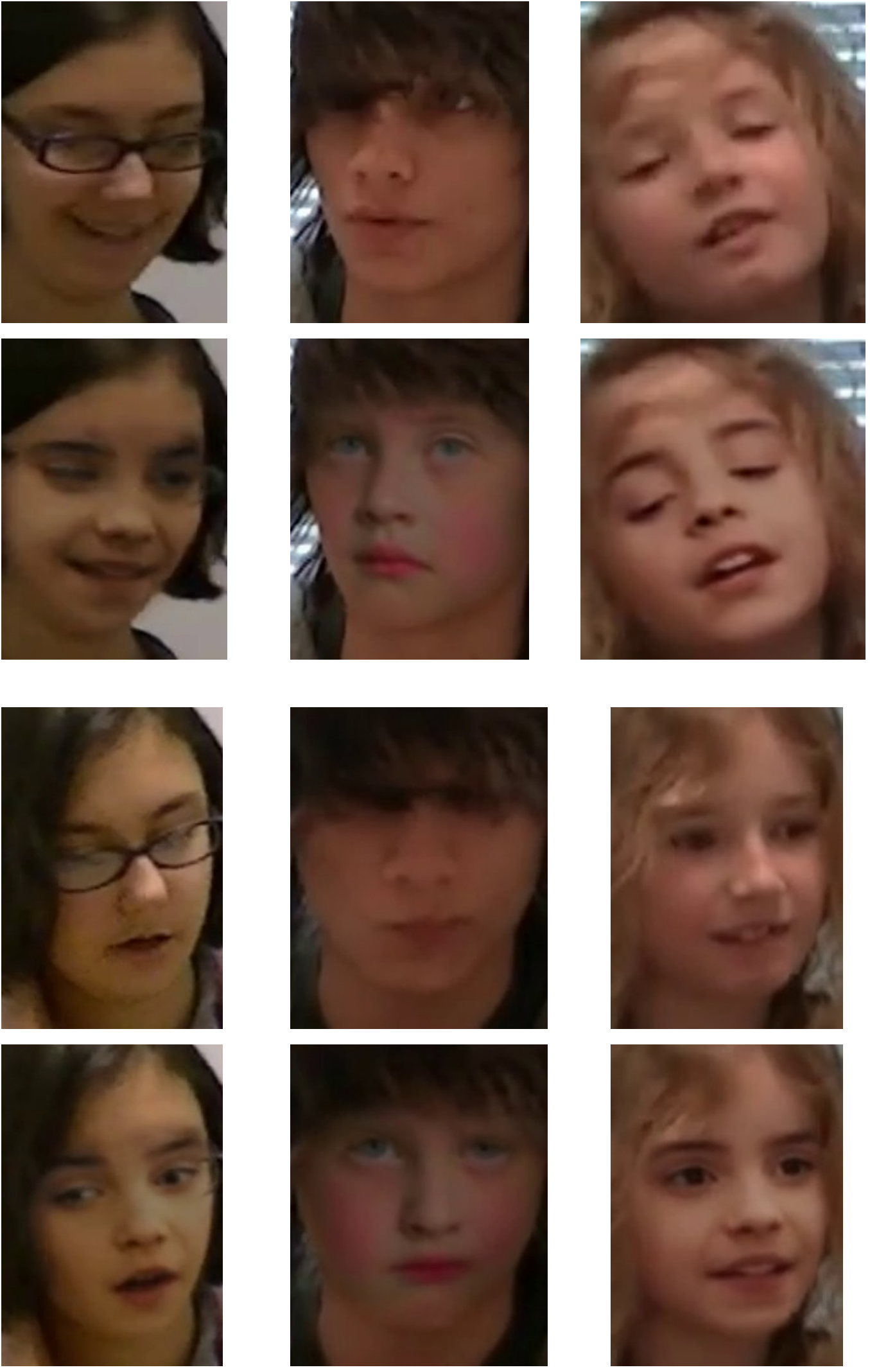}

    \caption{Additional visual pairings between original faces (top) and the face swaps produced by our system (bottom).}
\end{figure}

\end{document}